# Mathematical Reasoning Enhanced LLM for Formula Derivation: A Case Study on Fiber NLI Modelling

Yao Zhang, Yuchen Song, Xiao Luo, Shengnan Li, Xiaotian Jiang, Min Zhang, and Danshi Wang*

***Abstract*—Recent advances in large language models (LLMs) have demonstrated strong capabilities in code generation and text synthesis, yet their potential for symbolic physical reasoning in domain-specific scientific problems remains underexplored. We present a mathematical reasoning enhanced generative AI approach for optical communication formula derivation, focusing on the fiber nonlinear interference modelling. By guiding an LLM with structured prompts, we successfully reconstructed the known closed-form ISRS GN expressions and further derived a novel approximation tailored for multi-span C and C+L band transmissions. Numerical validations show that the LLM-derived model produces central-channel GSNRs nearly identical to baseline models, with mean absolute error across all channels and spans below 0.109 dB, demonstrating both physical consistency and practical accuracy.**

## I. Introduction

As optical networks scale in capacity and bandwidth, accurate physical layer modelling becomes critical for reliable performance evaluation and optimization. The Gaussian Noise (GN) model [1], balancing accuracy and computational complexity in estimating nonlinear interference (NLI) accumulation in fibers, has been widely adopted for quality of transmission (QoT) estimation across diverse transmission systems. To account for inter-channel simulated Raman scattering (ISRS) effects in multi-band wavelength division multiplexing (WDM) systems, several GN model extensions have been developed, including the generalized GN (GGN) model [2], enhanced GN (EGN) model [3], and ISRS GN model [4]. All of them typically relied on multidimensional numerical integration over frequency-related and span-dependent parameters. To improve computational efficiency, closed-form approximations of the EGN and ISRS GN models have also been proposed [5-7]. However, whether extending existing models or deriving closed-form expressions, the process remains heavily reliant on expert-driven mathematical derivation and scenario-specific assumptions. As optical networks evolve toward ultra-wideband operation, short-reach datacenter connection, high symbol rates, advanced modulation formats, and new-type fiber deployments, the manual derivation of models faces growing challenges in scalability and adaptability to the emerging scenarios, as depicted in Fig. 1. Therefore, this highlights the urgent need for more powerful tools to facilitate formula derivation and accelerate scientific research in optical communications.

The recent advancement of generative artificial intelligence (GenAI), particularly large language models (LLMs), has demonstrated impressive combinate applications in optical networks, including alarm/log analysis [8-9], performance optimization [10-12], autonomous configuration and control [13-15], and lifecycle management [16-18]. However, these preliminary applications mainly focused on general-purpose tasks involving data processing, textual understanding, and workflow automation. As LLMs continue to evolve, their potential to tackle fundamental scientific challenges has become increasingly prominent, with notable progress in chemical discovery [19], drug development [20], and material design, underscoring its role as a collaborative "co-scientist". Applying LLM to complex mathematical reasoning and physics-based modelling remains challenging, yet it holds exciting opportunities to advance scientific research in optical communications and beyond.

In this paper, we proposed a prompt-guided mathematical reasoning framework to empower LLM in autonomous formula derivation for physical modelling. First, we reconstructed the derivation of the closed-form ISRS GN model from its integral form using GPT-4o, significantly reducing reliance on manual efforts and obtaining consistent expressions with those presented in [4]. Moreover, based on the enhanced mathematical reasoning capability, we guided the LLM to discover a new approximate closed-form expression for the ISRS GN model applicable to wideband WDM systems. Numerical evaluations across C- and C+L-band transmissions demonstrate that the newly derived formula aligns closely with GGN and ISRS GN results, achieving good accuracy while streamlining the derivation process and minimizing expert dependence.

## II. Prompt-guided Derivation of Closed-Form ISRS GN Model

The ISRS GN model, essential for accurately assessing NLI effects, is widely used for wideband system design and performance evaluation. To explore the mathematical reasoning capability of GenAI models, we guided GPT-4o, an

This work was supported in part by the National Natural Science Foundation of China under Grants U24B20133 and 62171053, Beijing Nova Program 20230484331, and BUPT Excellent Ph.D. Students Foundation CX20241033. *(Corresponding author: Danshi Wang).*

Yao Zhang, Yuchen Song, Xiao Luo, Shengnan Li, Xiaotian Jiang, Min Zhang, and Danshi Wang are with the State Key Laboratory of Information Photonics and Optical Communications, Beijing University of Posts and Telecommunications, Beijing 100876, China (e-mail: zhang-yao@bupt.edu.cn; songyc@bupt.edu.cn; XiaoLuoD@bupt.edu.cn; shengnanli@bupt.edu.cn; jxt@bupt.edu.cn; mzhang@bupt.edu.cn; danshi_wang@bupt.edu.cn ).

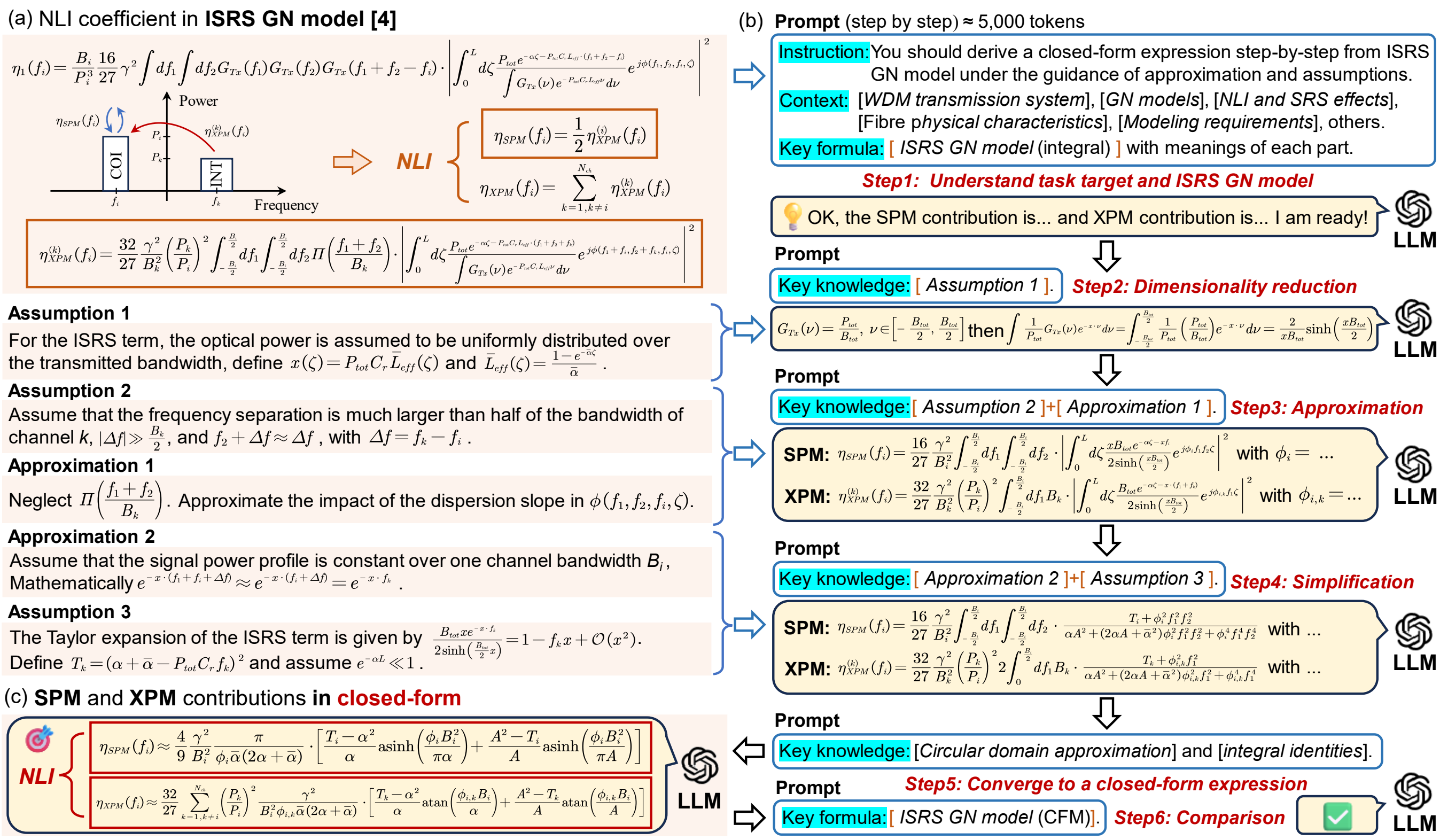

**Fig. 1.** The formula derivation process using GenAI with step-by-step prompt guidance of assumptions and approximations. (a) NLI coefficient in ISRS GN model. (b) The step-by-step assumptions and approximations provided through natural language prompts. (c) The closed-form expressions of SPM and XPM contributions derived by LLM consistently with ISRS GN model

advanced LLM known for its reasoning capabilities, to reconstruct the closed-form ISRS GN model from its integral form. Without access to any intermediate equations, the derivation was driven entirely by step-by-step modelling assumptions and approximations provided through natural language prompts.

The NLI effects on the channel of interest (COI) $i$ in ISRS GN model arise from two main sources: self-phase modulation (SPM) contribution $\eta_{SPM}(f_i)$ of the COI and cumulative cross-phase modulation (XPM) contributions $\eta_{XPM}^{(k)}(f_i)$ from all the interferer (INT) channels $k$. The key integral expressions for SPM and XPM (Eq. (8) and (9) in [4]), along with the physical interpretation of each term, were provided to the LLM as fundamental references, as illustrated in Fig. 1 (a). To facilitate efficient understanding of the derivation target, we compiled a comprehensive background for the LLM covering WDM transmission systems, existing GN models, NLI and ISRS effects, fiber physical characteristics, and modelling requirements. This context, amounting to approximately 5,000 tokens, was embedded into the prompts as essential knowledge to support physically consistent derivation.

Subsequently, we guided the LLM through a step-by-step derivation process, as shown in Fig. 1 (b), in which each prompt specified the relevant physical assumptions, such as uniformly distributed power profiles, frequency separation, Taylor expansion, and scenario-appropriate approximations. Following these structured prompts, the LLM independently performed dimensionality reduction, approximation, and simplification. Then, it separately converged the SPM and XPM contributions into a closed-form expression, consistent with Eq. (10) and (11) in [4], as in Fig. 1 (c). Notably, rather than recalling or reproducing memorized formulas, the LLM progressively reconstructed the derivation solely based on the supplied conditions.

## III. Newly Derived Closed-Form Expression Using LLM

Building upon the multiple extensions of GN model, we employed a five-step process to guide the LLM in deriving a new closed-form ISRS GN expression for accurate NLI estimation under wideband ISRS conditions, as illustrated in Fig. 2.

Step1, an approximate 8,000-token domain-rich prompt was constructed to provide the LLM with detailed background, including the fundamental GN theory, extensions of GN model, ISRS formulations, and the definitions of SPM and XPM terms. This enabled the model to analyze and learn the structure and assumptions of existing formulations.

Step2, based on the acquired knowledge, the LLM was directed to attempt new derivations of the SPM and XPM integrals. Explicit hints on integral limits, variable substitutions, and series expansion strategies were provided to facilitate convergence toward closed-form expressions without over-simplifying key dependencies.

Step3, the candidate formulas were further optimized through algebraic simplification and structural adjustments. Multiple alternatives were retained at this step, reflecting the LLM's ability to generate diverse forms of approximations.

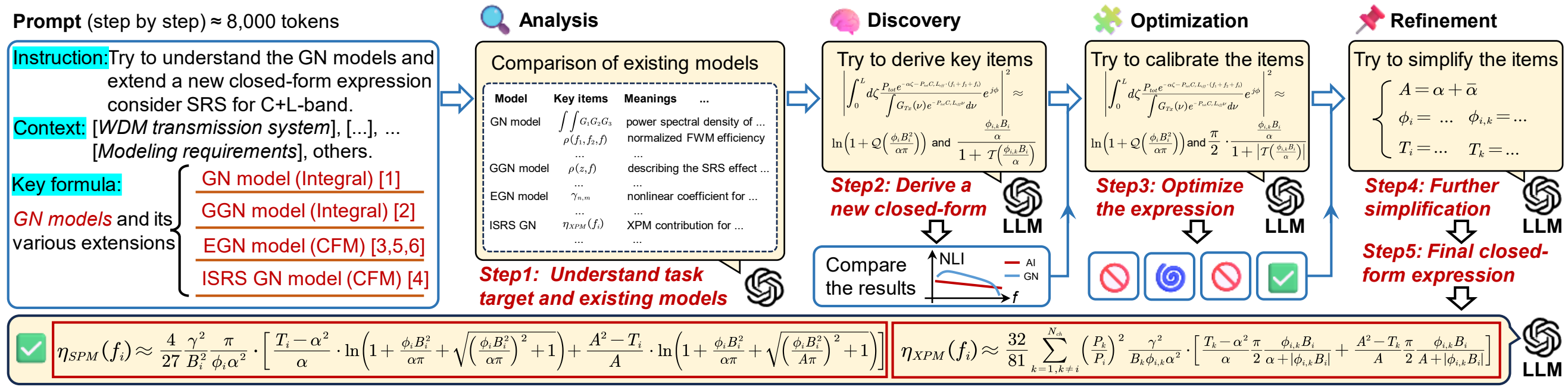

**Fig. 2.** Derivation process and the new closed-form ISRS GN formula driven by LLM

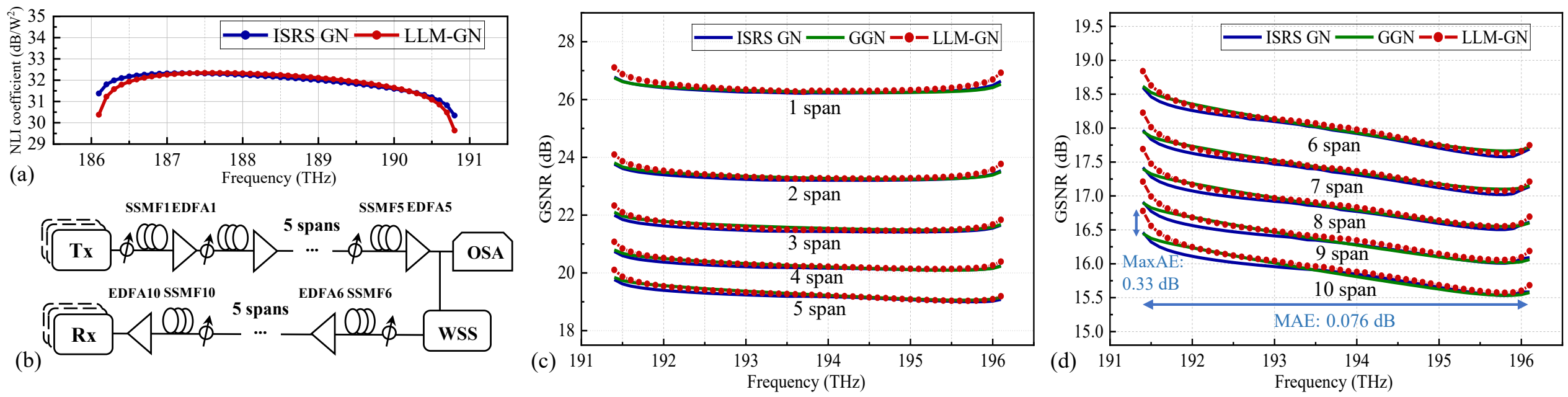

**Fig. 3.** Comparisons of the LLM-GN model in a C-band 10-span link. (a) NLI coefficient after single 100 km span using SSMF. (b) Simulation setup. (c) and (d) GSNR estimation by ISRS GN, GGN, and LLM-GN after each span.

Step4, iterative refinements were performed by comparing the LLM-derived expressions with established GN and ISRS GN models. Discrepancies in NLI and SRS contributions were analyzed and corrected, leading to progressively improved alignment with reference calculations across span counts and spectral loadings.

Step5, the LLM consolidated the optimized SPM and XPM components into a single, compact, new closed-form ISRS GN expression, which served as the outcome of the derivation process. This final formula retained clear physical interpretability, underscoring the LLM's capability for physics-guided symbolic derivation beyond rote reproduction of known equations.

Then, to evaluate the accuracy of the new LLM-derived formula (referred to as the LLM-GN model), we compared the NLI coefficient and generalized signal-to-noise ratio (GSNR) estimations after multi-span transmission using LLM-GN, GGN (by GNPy [21]), and ISRS GN [22] models across three scenarios.

*A. Full C-band transmission over 1 to 10 spans*

As shown in Fig. 3 (a), we first compared the NLI coefficient after a single 100 km span using standard single-mode fiber (SSMF). The LLM-GN model's predictions align closely with ISRS GN model, with only deviations on the edge of the C-band. Then, with 48 channels spaced at 100 GHz covering the C-band from 191.4 THz to 196.1 THz, we evaluated LLM-GN model in a 10-span link, where a wavelength selective switch (WSS) was inserted after the fifth span, as Fig. 3 (b). A uniform optimum launch power [23] for signals was used and Erbium-doped fiber amplifiers (EDFAs) with a fixed gain of 20.5 dB was set after each span to fully compensate for the fiber loss after 100 km transmission and a 0.5 dB insertion loss. The GSNR profiles predicted by LLM-GN model closely track those of reference models, exhibiting a mean absolute error (MAE) of 0.076 dB, and a maximum absolute error (MaxAE) of 0.33 dB at lower frequencies, as shown in Fig. 3 (c) and (d).

*B. Full C+L-band transmission over 1 to 10 spans*

To verify the scalability of the LLM-GN model in wideband systems, we also conducted a single-span NLI coefficient comparison for the C+L-band system after a single 100 km span SSMF, as in Fig, 4 (a). Then, we extended the evaluations to a C+L-band scenario, as shown in Fig. 4 (b), additional 48 channels were added covering the L-band from 186.1 THz to 190.8 THz, forming a 96-channel WDM system. Separate optimal launch powers were used for the C and L bands, while all other parameters were similar with the C-band case. Fig. 5 (a) and (b) show that the LLM-GN predicted GSNR profiles remain closely aligned with benchmarks throughout the entire link, particularly within the central 8 THz region in the first five spans, and 6 THz region after five spans. MAE is 0.092 dB across all spans and C+L-band channels, and MaxAE is 0.50 dB at the edge of the spectrum, confirming that the LLM-GN model achieves high-fidelity predictions both at span-level and for extended wideband links.

*C. C+L-band transmission with randomly loading conditions*

To further assess the robustness and generalization capability of the LLM-GN model, we evaluated its performance under non-uniform channel loading conditions. A total of 60 channels were used, with 32 randomly distributed across the C-band and 28 across the L-band, constructing a spectrally sparse WDM transmission. All the parameters matched those of the fully loaded C+L-band case. As depicted in Fig. 5 (c),

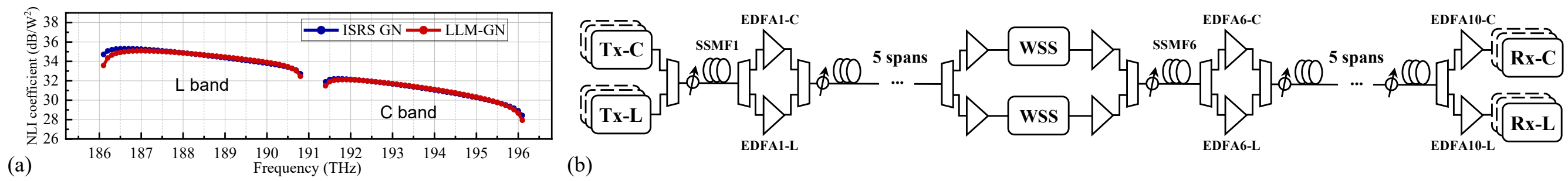

**Fig. 4.** Comparisons of the LLM-GN model in a C+L-band 10-span link. (a) NLI coefficient after a single span. (b) C+L-band simulation setup.

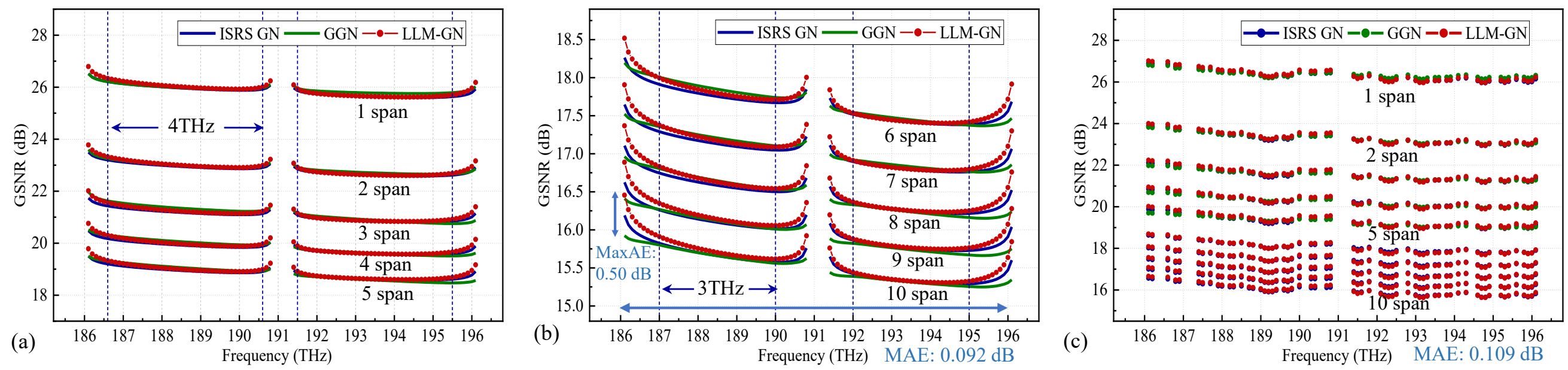

**Fig. 5.** Comparisons of the LLM-GN model in a C+L-band 10-span link. (a) and (b) GSNR estimation by ISRS GN, GGN, and LLM-GN after each span. (c) GSNR estimation with random 60-channels

the GSNR profiles predicted by the LLM-GN model remain in close consistent with reference models across all spans, despite the irregular spectral occupancy. Quantitatively, the MAE across all channels and spans is 0.109 dB, and MaxAE is 0.73 dB. The accuracy remains stable across the link, with no significant error accumulation observed in later spans. These results further confirm the validity of the LLM-GN model under diverse channel loading conditions.

## V. Conclusion

This work presents a promising application of GenAI for formula derivation, demonstrating its ability to reconstruct and extend physical models in optical communications through enhanced mathematical reasoning. By guiding the LLM in the derivation of the ISRS GN model, we not only reproduced known closed-form expressions but also obtained a new approximation that reduces reliance on manual derivations and expert heuristics. Numerical evaluations confirm that the proposed LLM-GN model achieves high accuracy across C-band, C+L-band, and dynamically loaded scenarios, with performance closely matching existing simulation tools. Beyond this specific case study, the results highlight the potential of GenAI to accelerate scientific discovery by enabling more autonomous, interpretable, and efficient physical-layer modelling beyond traditional human-centered workflows.